\documentclass[letterpaper, 10 pt, conference]{ieeeconf} 
\IEEEoverridecommandlockouts
\pdfoutput=1
\usepackage{graphics} 
\usepackage{epsfig} 
\usepackage{amsmath} 
\usepackage{amssymb} 
\usepackage{color}
\usepackage{bm}
\usepackage{subcaption}
\usepackage{multirow}
\usepackage{array}
\usepackage{cite}

\graphicspath{{figures/}}

\title{\LARGE \bf LiDAR-enhanced Structure-from-Motion}

\author{Weikun Zhen \quad 
        Yaoyu Hu \quad
        Huai Yu \quad
        Sebastian Scherer\thanks{Weikun Zhen is with the Department of Mechanical Engineering, Yaoyu Hu, Huai Yu and Sebastian Scherer are with the Robotics Institute. All authors are with Carnegie Mellon University, Pittsburgh, PA 15213. {\tt\small weikunz, yaoyuh, huaiy, basti@andrew.cmu.edu}}%
}
\begin{document}
\maketitle

​​\begin{abstract}
Although Structure-from-Motion (SfM) as a maturing technique has been widely used in many applications, state-of-the-art SfM algorithms are still not robust enough in certain situations. For example, images for inspection purposes are often taken in close distance to obtain detailed textures, which will result in less overlap between images and thus decrease the accuracy of estimated motion. In this paper, we propose a LiDAR-enhanced SfM pipeline that jointly processes data from a rotating LiDAR and a stereo camera pair to estimate sensor motions. We show that incorporating LiDAR helps to effectively reject falsely matched images and significantly improve the model consistency in large-scale environments. Experiments are conducted in different environments to test the performance of the proposed pipeline and comparison results with the state-of-the-art SfM algorithms are reported. 
\end{abstract} 
​\section{Introduction}
There is a growing demand for robot-based inspection that requires registering high-resolution image data of large scale civil engineering facilities, such as bridges and buildings. Those applications often use high-resolution, narrow Field-Of-View (FOV) cameras and images are taken in close distance to the structure surface for richer visual details. These properties pose new challenges to standard SfM algorithms. First of all, most available global or incremental SfM pipelines are based on a single camera, therefore, do not recover scale directly. More importantly, due to the limited FOV, the overlapped area between adjacent images is reduced, which will result in a pose graph that is only locally connected and jeopardize the accuracy of estimated motions. This issue becomes more significant in large-scale environments.

To address the mentioned challenges, we propose a novel pipeline that extends the traditional SfM algorithm to work with stereo cameras and LiDAR sensors. This work is based on a simple idea that LiDAR's long-ranging capacity can be used to restrain the relative motions between images. More specifically, we first implemented a stereo SfM pipeline that computes the motions of the camera and estimates the 3D positions of visual features (the structures). Then the LiDAR point clouds and the visual features are joined in a single optimization function, which is solved iteratively to refine the camera motions and structures. In our pipeline, the LiDAR data enhance the SfM algorithm in two folds: 1) LiDAR point clouds are used to detect and reject invalid image matches, making the stereo SfM pipeline more robust to visual ambiguities; 2) LiDAR point clouds are combined with visual features in a joint optimization framework to reduce the motion drifts. With the two folds of enhancement, our pipeline can achieve more consistent and accurate motion estimation than the state-of-the-art SfM algorithms. 

The contribution of this work can be summarized as follows: 
1) We adapt the global SfM techniques to a stereo camera system to initialize the camera motions in true scale.
2) LiDAR data are used to reject invalid matches of images, further robustifying the pipeline. 
3) We extend our previously proposed joint optimization pipeline by considering the shared structures of the stereo camera and LiDAR, which improves the accuracy and consistency of built models.

\begin{figure} [t]
    \centering
    \includegraphics[trim=0 0 0 0, clip, width=0.95\linewidth]{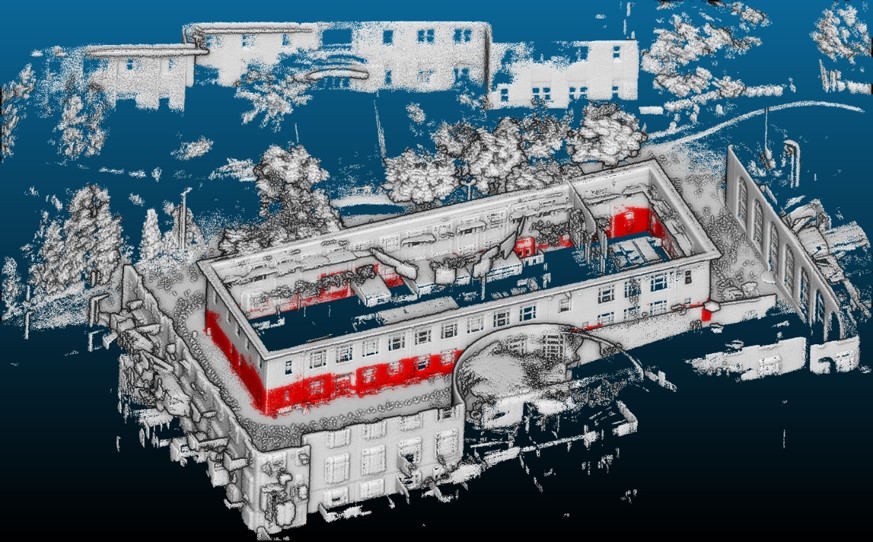}
    \caption{Reconstructed point cloud model (grey) of the CMU Smith Hall overlaid with visual feature points (red).}\vspace{-2mm}
    \label{fig:head}
\end{figure}

The rest of this paper is organized as follows: Section \ref{related} presents the related work on SfM and LiDAR-camera fusion techniques. Section \ref{sec:stereo_sfm} describes the proposed pipeline in detail. Experimental results are shown in Section \ref{sec:experiments}. Conclusions are discussed in Section \ref{sec:conclusion}.

\section{Related Work}
\label{related}
SfM has been an active research area over the last two decades. One of the most widely used open-source tools Bundler, also known as PhotoTourism \cite{snavely2006photo}, shows the ability to process a vast amount of online photos to build large-scale 3D models. Since then, significant advancements have been achieved in works such as Theia \cite{theia-manual}, VisualSFM \cite{wu2011visualsfm}, OpenMVG\cite{moulon2013global}\cite{moulon2012adaptive} and more recent COLMAP \cite{schonberger2016structure}. A comprehensive comparison of available SfM approaches is given in \cite{bianco2018evaluating} which concludes that OpenMVG and COLMAP achieved the state-of-the-art accuracy, robustness and completeness. Although many approaches exist, the majority are targeted on the general-purpose reconstruction, which does not deal with narrow FOV and close distance issues. Our work is focused on more detailed textures using narrow FOV cameras for inspection purposes. Additionally, a calibrated stereo pair is used together with a LiDAR sensor to provide metric information for the reconstructed model. 

Stereo pair is more popular in robotic applications to provide robot states, i.e. visual odometry (VO). As one of the most fundamental components of robot autonomy, many stereo-based VO algorithms have been proposed and can be roughly categorized into optimization-based approaches such as \cite{lupton2011visual} \cite{leutenegger2015keyframe} \cite{usenko2016direct}, and filter-based approaches as in \cite{paul2017comparative} and \cite{sun2018robust}. These applications typically use wide-angle cameras and require real-time performance. Differently, we focus on achieving higher robustness and model completeness by solving the reconstruction problem in a batch mode. Additionally, to deal with the problem of limited overlap between images, we relax the requirement that each feature must be observed by two stereo pairs (four views), which allows more valid features to be used for pose estimation. 

In addition to pure vision-based methods, people have also been fusing the depth information from LiDAR for robust and low-drift state estimation. A monocular camera combined with the LiDAR, as in \cite{zhang2015visual} \cite{graeter2018limo} \cite{shin2018direct}, is demonstrated to be able to provide real-time state estimation with true scale. In \cite{balazadegan2016visual} and \cite{shao2019stereo}, a stereo pair is used to robustify the state estimation. Although closely related, our work uses the LiDAR information to enhance components of a stereo SfM pipeline and globally optimizes the sensor poses. 

​\section{LiDAR-enhanced SfM}
\label{sec:stereo_sfm}
\begin{figure} [t]
	\centering
	\includegraphics[width=0.98\linewidth]{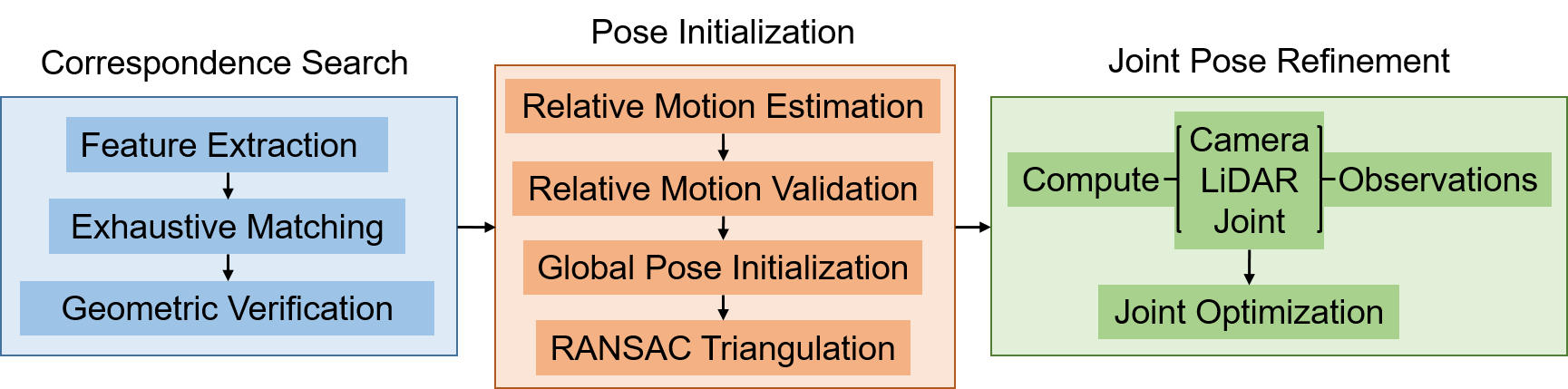}
	\caption{The LiDAR-enhanced stereo SfM pipeline.}\vspace{-2mm}
	\label{fig:pipeline}
\end{figure}
The proposed pipeline takes a set of stereo image pairs and associated LiDAR point clouds as inputs and generates 3D models of the covered environment in the format of triangulated visual points and a merged LiDAR point cloud. Fig. \ref{fig:pipeline} shows the procedures of our LiDAR-enhanced SfM pipeline, which is described in detail in this section. 

\subsection{Correspondence Search}
Given the stereo image pairs, computing the correspondences includes feature extraction, matching and geometric verification. Firstly, we rely on the OpenMVG library to extract SIFT \cite{lowe2004distinctive} features from images. Then the features are matched exhaustively (ignoring stereo pairs) using the provided cascade hashing method \cite{cheng2014fast}. Finally, The found matches between two images are verified by checking the two-view epipolar geometry. Specifically, the fundamental matrix $\bold F$ is estimated using RANSAC and then used to check the epipolar error of matched features. Only geometrically consistent features are kept for further computation. 
\begin{figure} [t]
    \centering
    \includegraphics[trim=0 0 0 0, clip, width=0.9\linewidth]{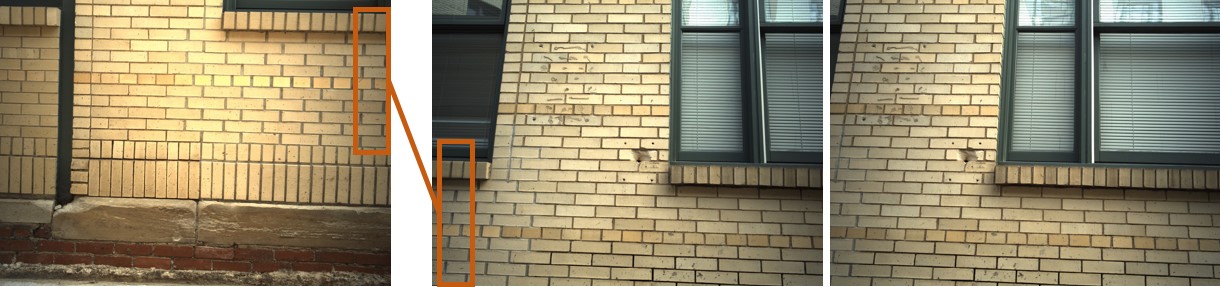}
    \caption{An example of regions for 2-view features. \textit{Left:} The right image of one station; \textit{Middle-Right:} The left and right images of the other station. The shared small regions are close to the boundries and marked by red boxes.}
    \label{fig:onelink}
\end{figure}
\subsection{Relative Motion Estimation} 
\begin{figure} [t]
	\centering
	\begin{subfigure}[b]{0.08\textwidth}
      \includegraphics[width=\textwidth]{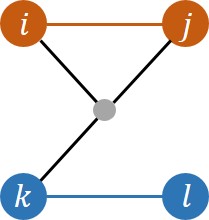}
      \subcaption{\textit{i-j-k}}
       \label{fig:link1a}
    \end{subfigure}
    \begin{subfigure}[b]{0.08\textwidth}
      \includegraphics[width=\textwidth]{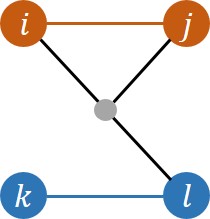}
      \subcaption{\textit{i-j-l}}    
       \label{fig:link1b}
    \end{subfigure}
    \begin{subfigure}[b]{0.08\textwidth}
      \includegraphics[width=\textwidth]{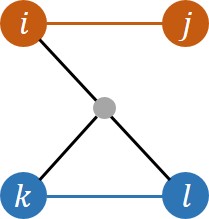}
      \subcaption{\textit{i-k-l}}
       \label{fig:link2a}
    \end{subfigure}
    \begin{subfigure}[b]{0.08\textwidth}
      \includegraphics[width=\textwidth]{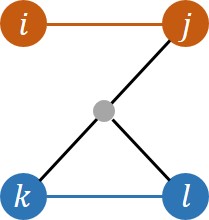}
      \subcaption{\textit{j-k-l}}    
       \label{fig:link2b}
    \end{subfigure}
    \begin{subfigure}[b]{0.08\textwidth}
      \includegraphics[width=\textwidth]{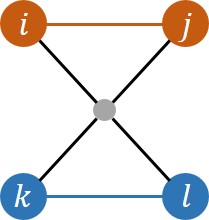}
      \subcaption{\textit{i-j-k-l}}    
       \label{fig:link2b}
    \end{subfigure}
    \caption{Examples of shared features (grey dot) between two stations (red and blue circle pairs). The colored bar indicates known stereo calibration. (a)-(d) 3-view; (e) 4-view.}
    \label{fig:links}
\end{figure}
Since the stereo pair is pre-calibrated, we treat a pair of left and right images as an independent unit, called a station. The pose of a station is defined as the pose of the left camera. To estimate the relative motion, a standard stereo method relies on feature points that are observed by all the four images in two stations, while we observe that many points are only shared by three or even two images. Ignoring those points could miss important information to estimate the camera motion, especially in the case that images have limited overlap. Therefore, we choose to explicitly handle different cases of shared views between two stations. 

Specifically, we consider feature points that are shared by at least 3 views to ensure metric reconstruction. Although points with only 2 views could help to estimate the rotation and the direction of translation, they are ignored here since those points typically come from small overlapped regions as shown in Fig. \ref{fig:onelink}. On the other hand, for 3-view or 4-view features, potential cases are shown in Fig. \ref{fig:links}. It is also possible that multiple types of shared features exist between two stations. To simplify the problem, we choose the type with most correspondences to solve the relative motion. In the case of 3-view, the points are first triangulated with the stereo pair and then the RANSAC+P3P algorithm is used to solve the transform. In the case of 4-view, we follow the standard treatment that first triangulates the points in the two stations and then applies RANSAC+PCA registration algorithm to find the relative motion. In both cases, a non-linear optimization procedure is used to refine the computed poses and triangulations by minimizing the reprojection error of inliers. Finally, all poses are transformed to represent the relative motion between left cameras.

\subsection{Relative Motion Validation}
\begin{figure} [t]
    \centering
    \begin{subfigure}[b]{0.45\linewidth}
      \includegraphics[trim=0 0 0 0, clip, width=\linewidth]{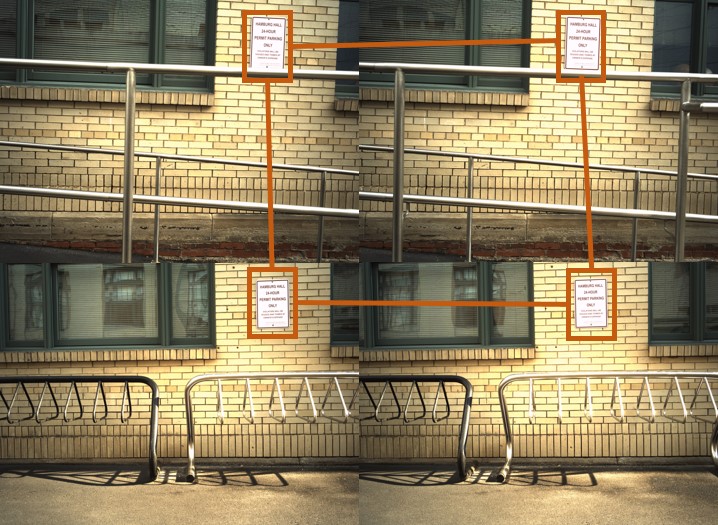}
      \subcaption{}
       \label{fig:lidar_valid_1}
    \end{subfigure}
    \begin{subfigure}[b]{0.485\linewidth}
      \includegraphics[trim=0 0 0 0, clip, width=\linewidth]{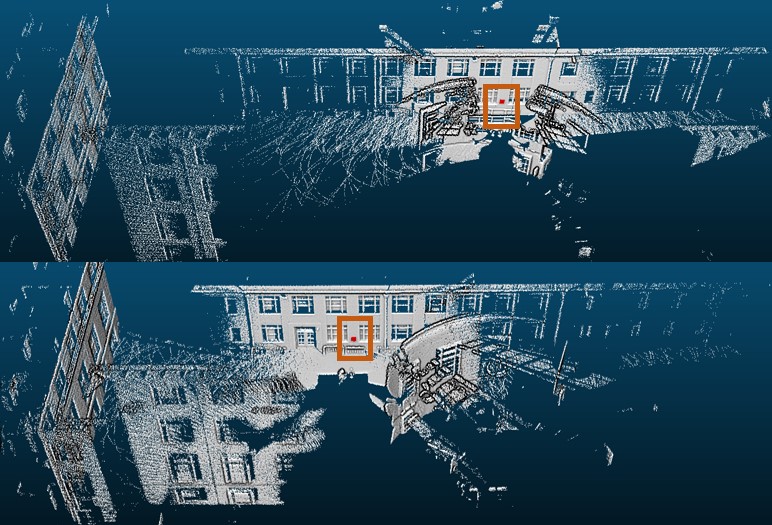}
      \subcaption{}
       \label{fig:lidar_valid_2}
    \end{subfigure}\\
    \begin{subfigure}[b]{0.98\linewidth}
      \includegraphics[trim=0 0 0 0, clip, width=\linewidth]{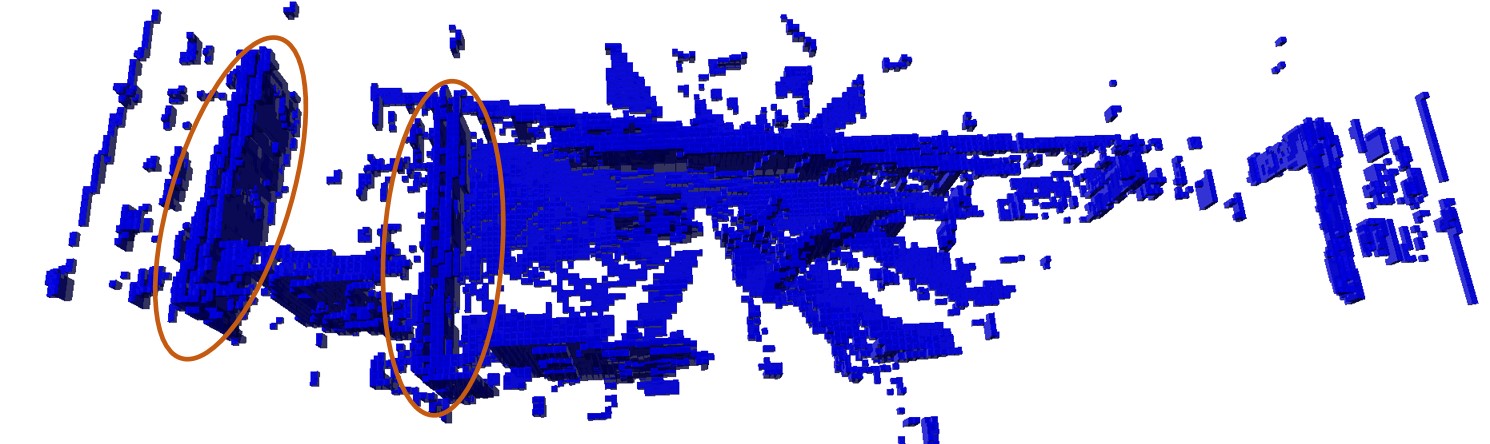}
      \subcaption{}
       \label{fig:lidar_valid_3}
    \end{subfigure}
    \caption{An example of invalid relative motion due to visual ambiguities. (a) Two pairs of images are matched incorrectly due to the identical parking sign. (b) The corresponding point clouds from the two stations with the signs marked by red boxes. (c) The merged occupancy grids showing incorrect alignment (red ellipses). The consistency ratio, in this case, is 0.56 while it is typically over 0.7 for valid relative motions. }\vspace{-2mm}
    \label{fig:validation}
\end{figure}
Once the relative motions are found, a pose graph can be built with the nodes representing the station poses and edges representing the relative motions. The global poses can then be solved by averaging the relative motions on the pose graph. However, it is likely that invalid edges exist due to visual ambiguities in the environment (see Fig. \ref{fig:lidar_valid_1} and \ref{fig:lidar_valid_2}) and directly averaging relative motions could give incorrect global poses. Therefore, a two-step edge validation scheme is designed to remove outliers. 

\begin{figure} [t]
    \centering
     \begin{subfigure}[b]{0.6\linewidth}
      \includegraphics[trim=0 0 0 0, clip, width=\linewidth]{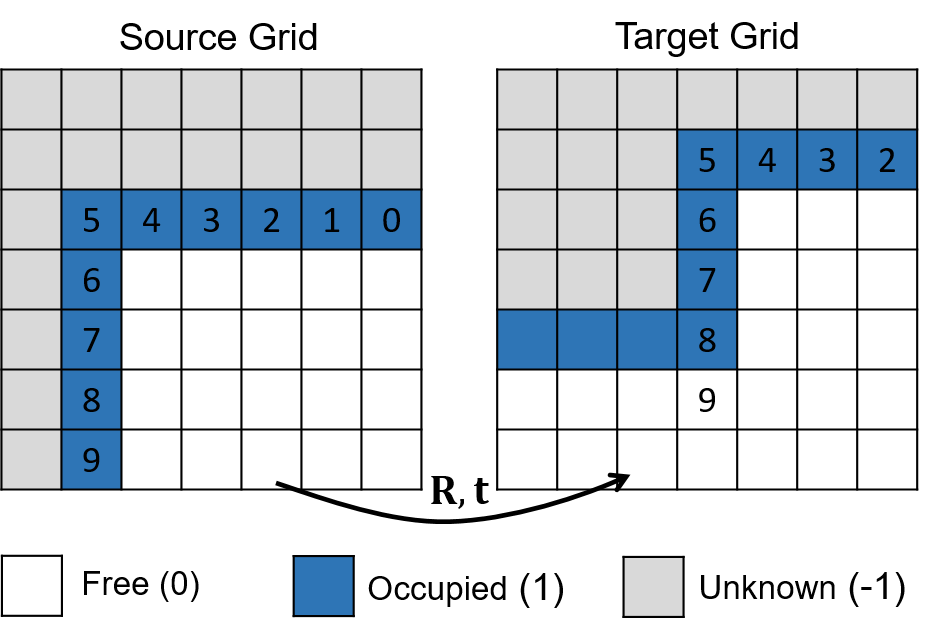}
      \subcaption{}
       \label{fig:grid}
    \end{subfigure}
    \begin{subfigure}[b]{0.25\linewidth}
      \includegraphics[trim=0 0 0 0, clip, width=\linewidth]{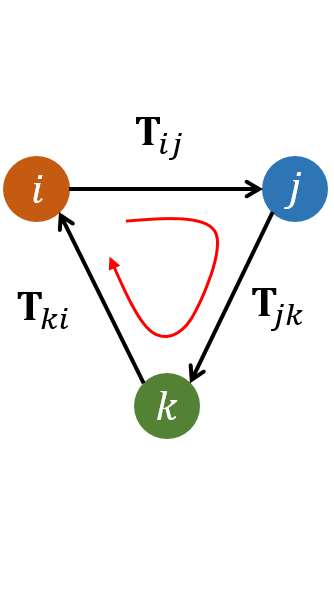}
      \subcaption{}
       \label{fig:triplet}
    \end{subfigure}
    \caption{(a) A 2D example of checking the consistency of occupancy grids. The occupied cells (labeled as 0-9, painted in blue) of the source grid are transformed into the target grid using $(\bold R, \bold t)$. Cells 2-8 are confirmed by the target grid while cell 9 is inconsistent. Therefore the consistency ratio is computed as $7/8$. (b) Transform-based cycle consistency checking. }
    \label{fig:validation}
\end{figure}

In the first step, we check the overlap of LiDAR point clouds for all station pairs and reject inconsistent ones. Specifically, two occupancy grids, namely the source grid $\mathcal{G}_s$ and target grid $\mathcal{G}_t$, are constructed from the corresponding point clouds. Each cell in the grids takes the value 1 if occupied, 0 if free, and $-1$ if unknown. Then for each occupied cell in $\mathcal{G}_s$, we transform the its center $\bold c\in \mathbb{R}^3$ to the frame of $\mathcal{G}_t$ using the estimated relative motion $(\bold R, \bold t)\in(SO(3), \mathbb{R}^3)$, and check the cell state at that location. In order to quantify the consistency, we define a consistency ratio as
\begin{align}
r_c = \dfrac{\sum_{\bold c \in \{\mathcal{G}_s(\bold c) = 1\}} I\left(\mathcal{G}_t(\bold R \bold c + \bold t) = 1\right)}{\sum_{\bold c \in \{\mathcal{G}_s(\bold c) = 1\}} I\left(\mathcal{G}_t(\bold R \bold c + \bold t) \neq -1\right)}
\end{align}
where $I(\cdot)$ is the indicator function that equals 1 if the argument is true, or 0 otherwise. The consistency ratio measures how much the $\mathcal{G}_t$ agrees with $\mathcal{G}_s$. In addition, we perform cross-checking by switching the target and source grids and recompute the consistency ratio. A relative motion is treated as valid only if both the ratios are larger than 0.6 (see Fig. \ref{fig:lidar_valid_3} for an example of inconsistent grids). Note this is a loose threshold since we encourage more pairs to remain for completeness and rely on further checks to reject remaining outliers. Fig. \ref{fig:grid} shows a 2D example of checking the consistency of two occupancy grids. In the implementation, the Octomap library \cite{hornung2013octomap} is used to efficiently iterate through occupied cells and look up the occupancy at a query location. 
 
As the second step of validation, we check the cycle consistency as in \cite{openMVG}. Specifically, we check small cycles of length 3, namely triplets. Here small cycles are preferable since they are easy to find and check. And more importantly, small cycles capture local information hence will not be affected by accumulated drifts. Note that the triplet here is a collection of 3 stations instead of images. As shown in Fig. \ref{fig:triplet}, we composite the relative motions $\bold T_{ki}\bold T_{jk}\bold T_{ij} \in {SE}(3)$ and check the resulting angle and translation. $2^\circ$ and $0.1$m are used as thresholds to label a triplet check as pass or not. For each relative motion, we keep track of its success rate $r_s$ defined as 
\begin{equation}
r_s = \frac{\text{\# of passed checks}}{\text{\# of involved checks}}
\end{equation}
and those with $r_s$ lower than 0.6 are treated as invalid. This threshold will be increased if invalid motions still exist. 



\subsection{Global Pose Initialization}
\label{sec:global_poses}
Given the set of valid relative motions $\mathcal{E} = \{\bold T_{ij}\in SE(3)\}$, initializing the global poses $\mathcal{T}=\{\bold T_i\in SE(3)\}$ can be easily achieved by solving the pose graph $(\mathcal{T}, \mathcal{E})$ with relative motions being the constrains. We first construct a maximum spanning tree (MST) using $\mathcal{E}$ which is weighted by the total number of feature correspondences used to estimate $\bold T_{ij}$. Then $\mathcal{T}$ can be constructed using edges from the MST. Finally, all pairs in $\mathcal{E}$ are considered, and the following cost function is used to optimize the poses in a global frame:
\begin{align}
\mathcal{T} & = \arg \min\sum_{\bold T_{ij}\in \mathcal{E}} || \log(\bold{T}_{ij}^{-1}\bold{T}_i^{-1}\bold{T}_j)||^2 
\end{align}
where the function $\log(\cdot): SE(3)\rightarrow \mathbb{R}^6$ computes the twist vector of a rigid body transform. 
\subsection{RANSAC Triangulation} 
\label{sec:triangulation}
We adopt the robust triangulation approach proposed in \cite{schonberger2016structure} which uses RANSAC for each 3D feature point to find the best views for triangulation. For each track, which is a collection of observations of one feature in different camera views, two views are sampled randomly and the DLT method \cite{hartley2003multiple} is used to triangulate the point. By projecting the point onto other views and selecting the ones with a small reprojection error, inlier views can be found. This process is repeated for a number of times and the largest set of inlier views (minimum 3 views are required) is retained. Finally, inlier views are used to refine the position of a feature point in the global frame by minimizing the reprojection error. 
%
%
\subsection{Joint Pose Refinement}

The pose refinement for vision-based SfM algorithms is typically achieved by Bundle Adjustment (BA). However, due to several system reasons, such as inaccurate feature locations, calibration inaccuracy, correspondence outliers and so on, the pose estimation could have significant drifts over long distances, especially when loop closures can't be found effectively. To address this issue, we consider making use of LiDAR's long-ranging capacity to constrain the camera motion. 
Our previous work \cite{zhen2019joint} proposed an optimization pipeline that jointly minimizes over the camera and LiDAR observations. In this work, we further introduce the joint observations that access the shared structures of the camera and the LiDAR. 

\subsubsection{Camera Observations}

Camera observations are defined as the feature coordinates on images and are directly obtained as inliers from the triangulation step. We define the camera observation error $\bold e_c$ as the feature reprojection error, namely
\begin{align}
\bold e_c = \pi(\bold x; \bold T_i, \bold P) - \bold u
\end{align}
where function $\pi$ projects feature point $\bold x\in\mathbb{R}^3$ onto the image plane, $\bold P \in \mathbb{R}^{3\times 4}$ is the camera projection matrix to be chosen depending on $\bold x$ is observed by the left or right camera, $\bold u \in \mathbb{R}^2$ is the observed image coordinates. We use the symbol $\mathcal{C}$ to denote the camera observation set and $\mathcal{S}$ as the set of reconstructed feature points in 3D.

\subsubsection{LiDAR Observations}
A LiDAR observation is defined as a matched pair of a key point $\bold p\in\mathbb{R}^3$ in station $i$ and a local patch $(\bold y, \bold n)\in(\mathbb{R}^3, S(2))$ in station $j$. Here $\bold y$ is the nearest neighbor of $\bold p$ in station $j$ and $\bold n$ is the normal vector of the local patch. The LiDAR observation error $\bold e_l$ is defined as the point to patch distance:
\begin{align}
\bold e_l = \bold n^T\left(\tau(\bold p; \bold T_e^{-1}\bold T_j^{-1}\bold T_i \bold T_e) - \bold y\right)
\end{align}
where function $\tau(\cdot)$ transforms a point based on the given transformation matrix, $\bold T_e$ is the extrinsic transform of LiDAR w.r.t. the left camera. Our algorithm relies on a reasonable initial guess of $\bold T_e$, but will also adjust it along with the poses and structures in the optimization step. Finally, we sample 5000 key points from each point cloud and search for the corresponding local patches from stations that are within a 5-meter range. The constructed LiDAR observation set is denoted by symbol $\mathcal{L}$.

\subsubsection{Joint Observations}
Similar to a LiDAR observation, a joint observation is also defined as a point-patch pair, except the point here comes from $\mathcal{S}$ instead of a LiDAR point cloud. For each feature point $\bold x \in\mathcal{S}$, we iterate through its views to find the point clouds potentially observing the same structure. Then for each point cloud, $\bold x$ is matched to the nearest local patch $(\bold y, \bold n)$ which shares the same definition as in the LiDAR observations. The joint observation error $\bold e_j$ is defined as
\begin{align}
\bold e_{j} = \bold n^T\left(\tau(\bold x; \bold T_e^{-1}\bold T_i^{-1}) - \bold y\right), 
\end{align}
and the symbol $\mathcal{J}$ is used to denote the set of extracted joint observations.

\subsubsection{Joint Optimization}
The final joint optimization is done by minimizing the previously defined observation errors in a single cost function:
\begin{align}
\arg \min \lambda_c\sum_{\mathcal{C}} \bold e_c^2 + \lambda_l\sum_{\mathcal{L}} \bold e_l^2 + \lambda_j\sum_{\mathcal{J}} \bold e_j^2
\label{eqn:total_cost}
\end{align}
to adjust camera poses $\mathcal{T}$, structures $\mathcal{S}$ and camera-LiDAR extrinsic transform $\bold T_e$ simultaneously. Here each component is weighted by a hand-tuned factor. In practice, we set $\lambda_c$ to be $1$ and tune the other two such that the cost of individual components are roughly of the same magnitude. Additionally, the error terms in (\ref{eqn:total_cost}) are wrapped with robust the Huber loss function and Ceres Solver \cite{ceres-solver} is used to solve the optimization problem.

Once the optimization procedure converges, observations are filtered by the errors and those with error larger than a threshold are removed from the observation set. $4$ pixels, $0.1$m and $0.1$m are used as thresholds for the camera, LiDAR and joint observations respectively. The cost function (\ref{eqn:total_cost}) is solved again once outliers are removed. 

Noticing that the LiDAR and joint observations are found through nearest neighbor search, which is highly dependent on the accuracy of the initialization. Therefore, the LiDAR and joint observations are recomputed once a new estimation of $\mathcal{T}$ and $\bold T_e$ is obtained. This process is repeated for several iterations until the cost difference diminishes. In our experience, 4-6 iterations are typically good enough to generate well-aligned models. 
​\section{Experiments}
\label{sec:experiments}

\begin{figure} [t]
    \centering
    \includegraphics[width=0.95\linewidth]{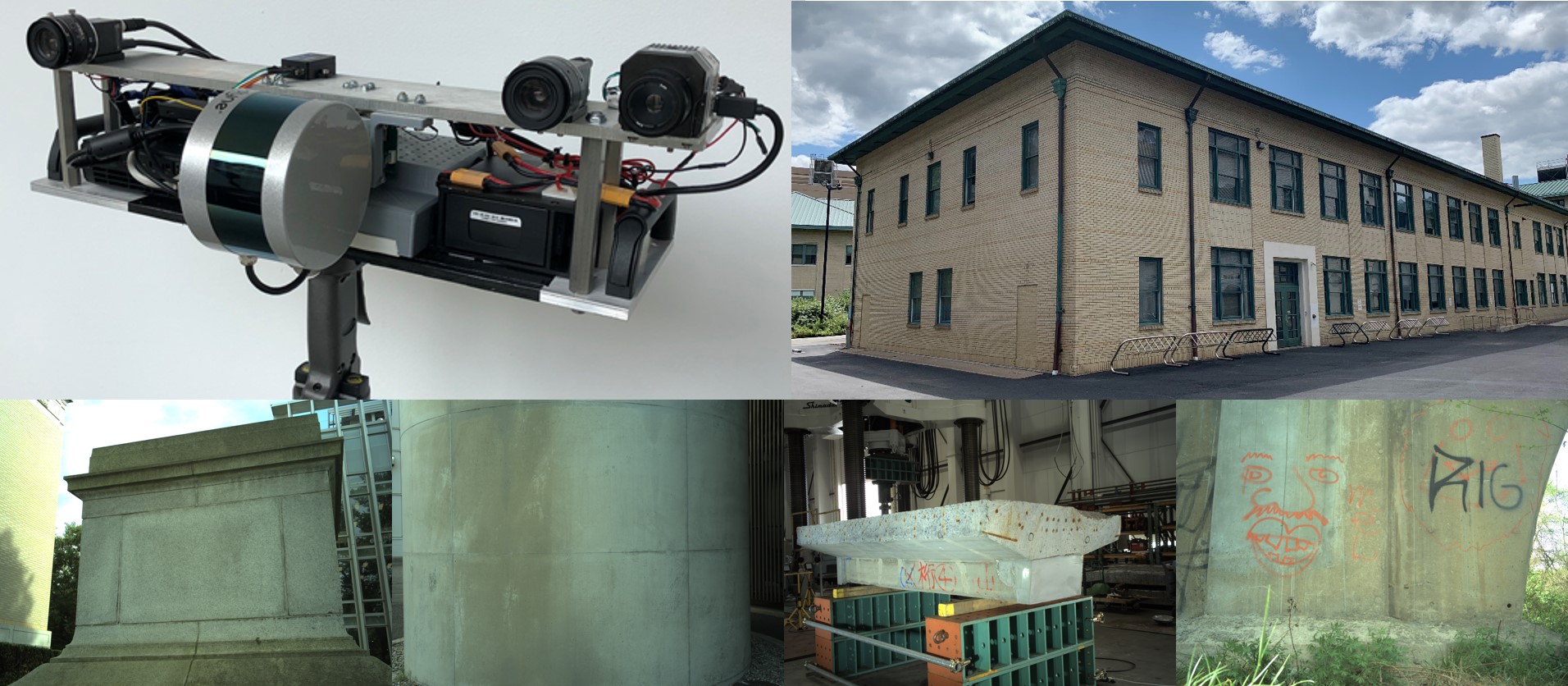}    
    \caption{The sensor pod and datasets. Top right is the CMU Smith Hall dataset. In the second row, from left to right are a concrete square pillar, a cylinder pillar, a T-shaped specimen, and a bridge pillar. }\vspace{-0mm}
    \label{fig:drone}
\end{figure} 
\subsection{Sensor Pod and Datasets}
A sensor pod is developed to have multiple onboard sensors, including two Ximea color camera (12 megapixel, global shutter), and a Velodyne Puck LiDAR (VLP-16) mounted on a continuously rotating motor. The scanned points from the VLP-16 are transformed into a fixed base frame using the motor angle measured by an encoder. 
Additionally, we assume the stereo camera and the LiDAR-motor system are both pre-calibrated. The LiDAR scanned points are transformed from the rotating LiDAR frame to a fixed motor frame, which is then referred to as the LiDAR frame. The extrinsic transform from the left camera to the fixed LiDAR frame is jointly optimized as $\bold T_e$ in our pipeline.

In this work, we collect our own datasets instead of using public SfM datasets since most of them are not suitable for our specific case that requires stereo cameras and a LiDAR. As shown in Fig. \ref{fig:drone}, our datasets vary from simple concrete structures to a large scale building. The Smith Hall dataset contains 276 stations (552 images and 276 LiDAR point clouds) and is most challenging due to repeated patterns and limited overlap between images. The rest datasets in the second row are of smaller scale and contain 29, 54, 25, 32 stations respectively.
\subsection{Relative Motion Estimation}
\begin{figure} [t]
    \centering
    \includegraphics[trim=30 0 30 0, clip, width=0.99\linewidth]{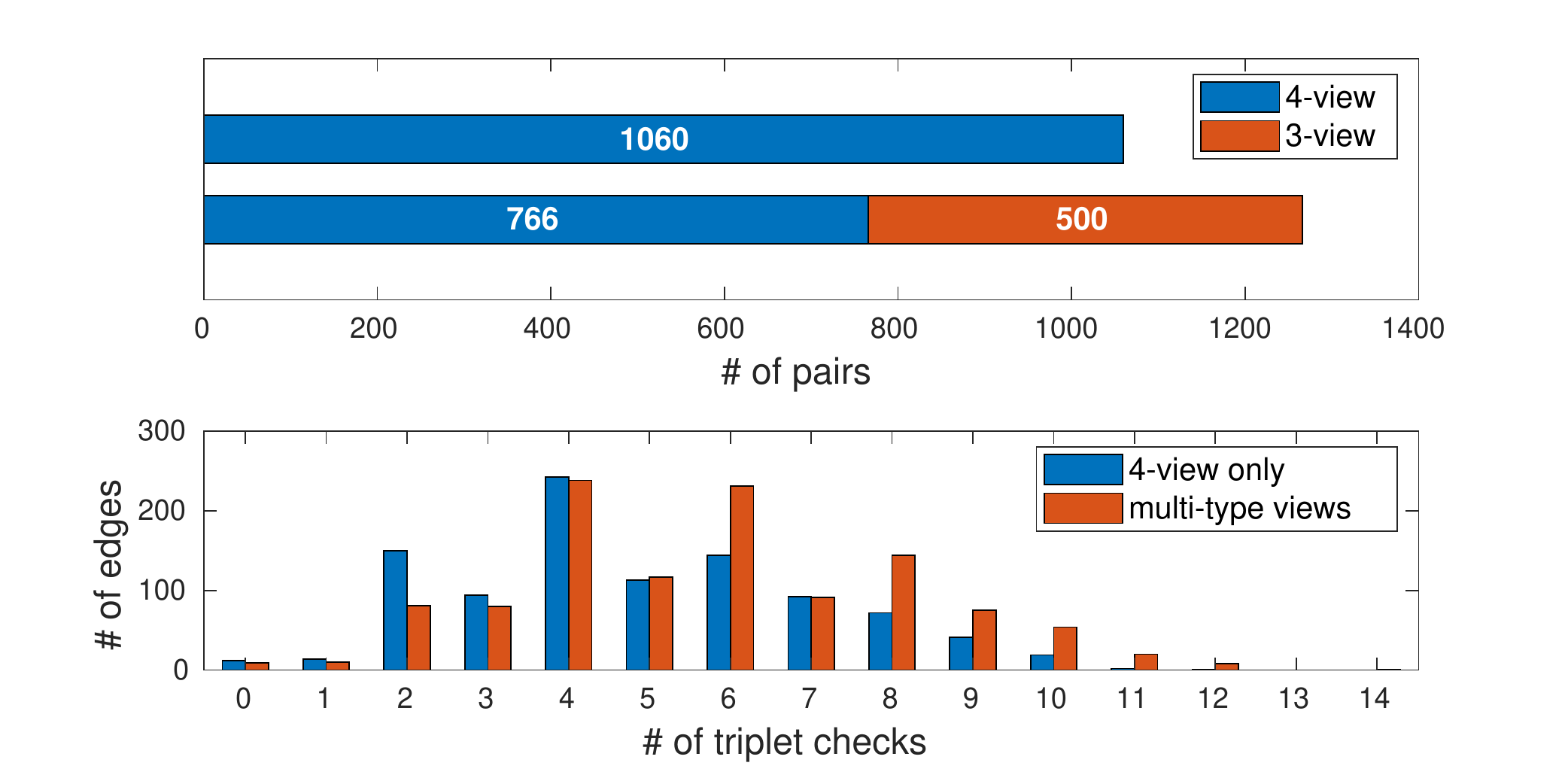}
    \caption{ \emph{Up:} Visualization of the number of the pairs solved from 4-view and 3-view points. \emph{Down:} Histograms of the number of edges for different number of triplet checks. 
}\vspace{-1mm}
    \label{fig:link_count}
\end{figure}

\begin{figure} [t]
    \centering
    \includegraphics[trim=0 0 0 0, clip, width=0.4\linewidth]{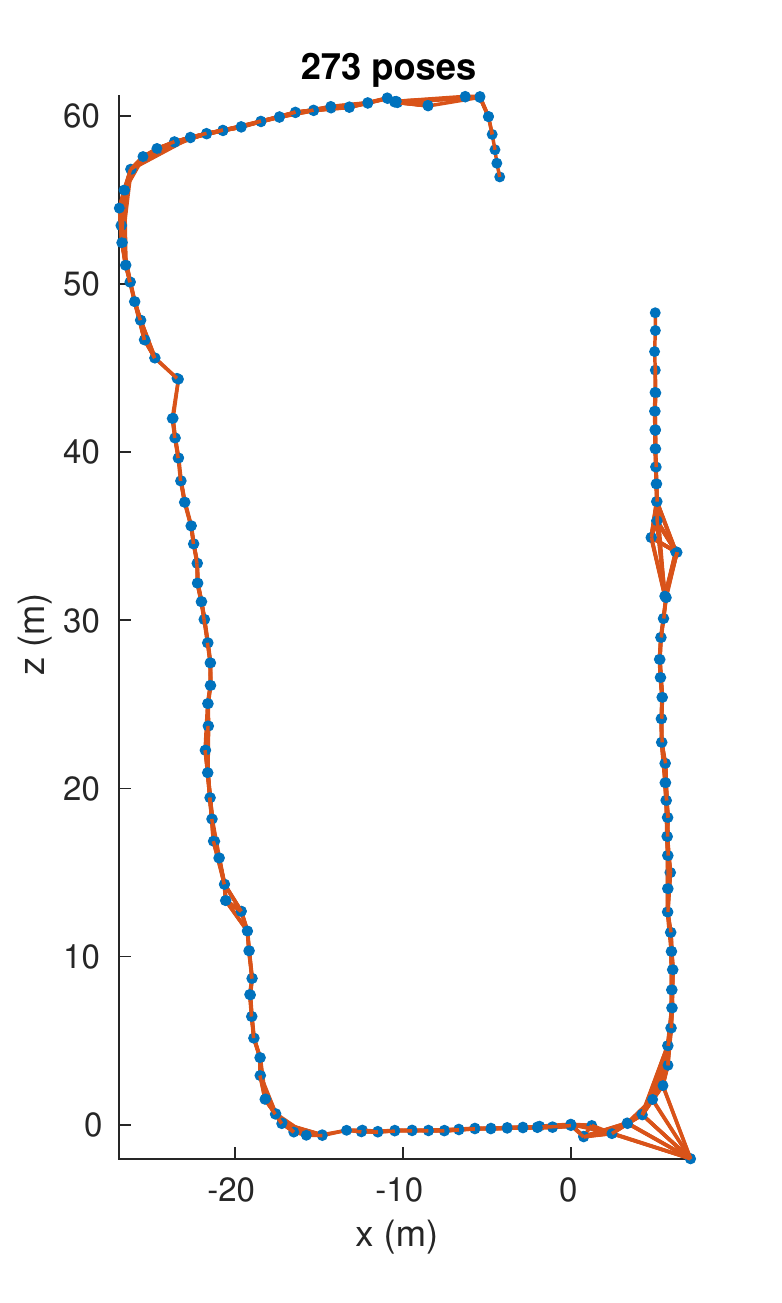}
    \includegraphics[trim=0 0 0 0, clip, width=0.4\linewidth]{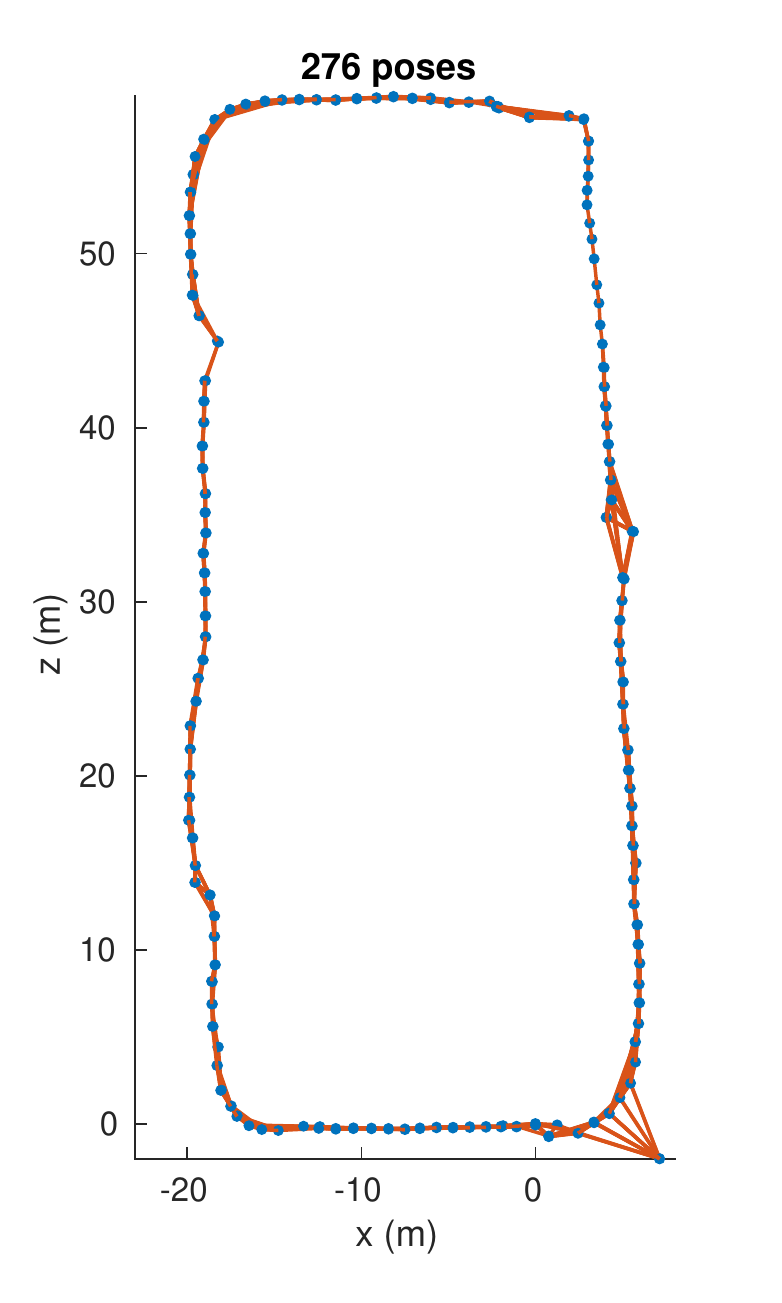}
    \caption{ \emph{Left:} Initialized pose graph with 4-view features. \emph{Right:} Initialized pose graph with multi-type views. }\vspace{-1mm}
    \label{fig:graph}
\end{figure}

In this experiment, we show that relaxing the 4-view requirement of a feature point allows recovering more valid relative motions, which is beneficial for cycle consistency checking and model completeness. Here the Smith Hall dataset is used. As shown in the upper plot of Fig. \ref{fig:link_count}, 500 edges are computed thanks to the 3-view feature points. Compared with the case of only 4-view features, 206 ($19.4\%$) more relative motions are recovered. The lower plot of Fig. \ref{fig:link_count} shows a histogram of the number of edges w.r.t. the number of involved triplets. We observe that the edges tend to be involved in more triplets, implying that more frequent checks can be conducted. The total number of triplet checks for this dataset is therefore increased from 4.8K to 6.7K. Finally, the initialized global poses are visualized in Fig. \ref{fig:graph}. Because more relative motions are recovered, our algorithm can connect more stations (276 vs. 273). Surprisingly, even 3 more connected stations can make the algorithm successfully pick up the loop closure around the top right corner of the model. 

\subsection{Relative Motion Validation}
In this section, we compare the outlier rejection performance of the proposed grid-based check (GC, with $r_c$ threshold being 0.6) and success rate check (SR), to rotation cycle check (RC) used by OpenMVG \cite{openMVG} and transform (rotation and translation) cycle check (TC). Again, the smith hall dataset is used.
\begin{table}[t]
\caption{A comparison of validation methods.}
\label{tab:confusion}
\centering
\setlength\tabcolsep{3pt}
\begin{tabular}{c||c|c|c}
\hline
        & Confusion Matrix & Recall & Precision \\
\hline
RC& [1151, 0; 91, 24] & 20.87\% & 100.00\%\\
\hline
TC& [1151, 0; 88, 27] & 23.48\% & 100.00\%\\
\hline
GC& [1149, 2; 10, 105] & 91.30\% & 98.13\%\\
\hline
SR(0.9)& [1035, 116; 5, 110] & 95.65\% & 48.67\%\\
\hline
SR(0.6)& [1139, 12; 40, 75] & 65.22\% & 86.21\%\\
\hline
GC+SR(0.9)& [1033, 118; 3, 112] & 97.39\% & 48.70\%\\
\hline
GC+SR(0.6)& [1137, 14; 6, 109] & 94.78\% & 88.62\%\\
\hline
\end{tabular}
\end{table}
\begin{figure*} [t]
    \centering
    \begin{subfigure}[b]{0.2\linewidth}
      \includegraphics[width=\linewidth]{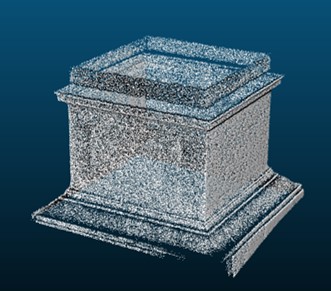}
      \subcaption{Square}
       \label{fig:smith_colmap}
    \end{subfigure}
    \begin{subfigure}[b]{0.221\linewidth}
      \includegraphics[width=\linewidth]{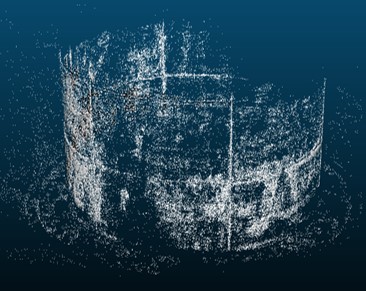}
      \subcaption{Cylinder}
       \label{fig:recon_cylinder}
    \end{subfigure}
    \begin{subfigure}[b]{0.267\linewidth}
      \includegraphics[width=\linewidth]{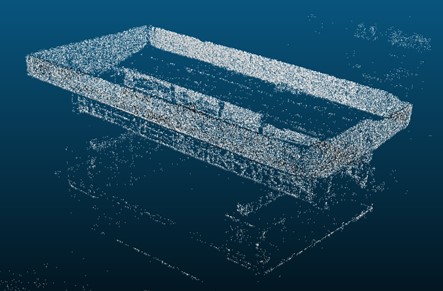}
      \subcaption{T-shaped}
       \label{fig:recon_t}
    \end{subfigure}
    \begin{subfigure}[b]{0.24\linewidth}
      \includegraphics[width=\linewidth]{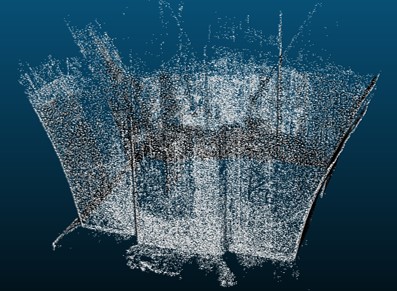}
      \subcaption{Bridge Pillar}
       \label{fig:recon_bridge}
    \end{subfigure}~\\
    \begin{subfigure}[b]{0.32\linewidth}
      \includegraphics[width=\linewidth]{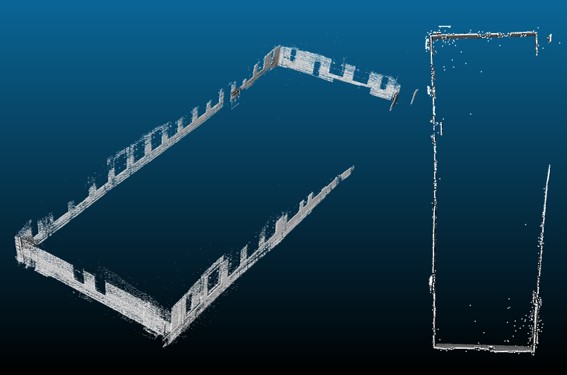}
      \subcaption{COLMAP}
       \label{fig:smith_colmap}
    \end{subfigure}
    \begin{subfigure}[b]{0.305\linewidth}
      \includegraphics[width=\linewidth]{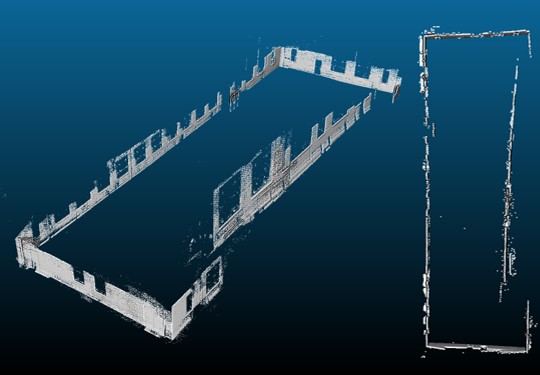}
      \subcaption{OpenMVG}
       \label{fig:smith_mvg}
    \end{subfigure}
    \begin{subfigure}[b]{0.313\linewidth}
      \includegraphics[width=\linewidth]{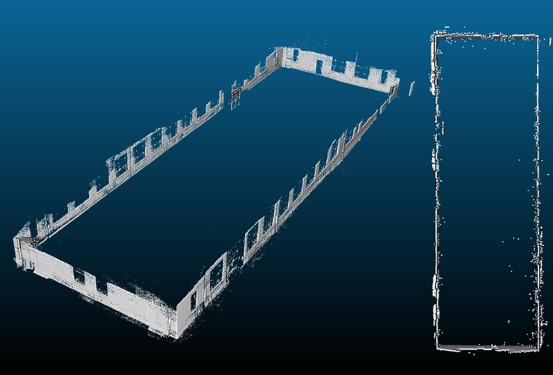}
      \subcaption{Ours}
       \label{fig:smith_joint}
    \end{subfigure}\\
    \caption{Reconstruction results on collected datasets.}
    \label{fig:smith_recon}
\end{figure*} 

Firstly, we obtain the ground truth label by checking the difference of each relative motion to the global poses produced by our pipeline, which is of reasonable accuracy as shown in Fig. \ref{fig:head}. Then those pairs with an angular error greater than $2^{\circ}$ or translational error greater than $0.1$m are labeled as outliers. After that, the validation methods mentioned above are used to check the given relative motions. Specifically, a confusion matrix is computed for each case and is shown in Table \ref{tab:confusion}. We use \emph{recall} to measure the ratio of picked true outliers among all true outliers and \emph{precision} to denote the fraction of true outliers among the predicted instances. We observe that a number of inconsistent relative motions could pass the RC or TC checks, corrupting the initialization of global poses. On the other hand, GC and SC checks apply more strict passing rules, hence more outliers can be rejected effectively. However, it is also observed that many valid motions fail to pass the SR check. The number is regulated by the threshold of $r_s$. In reconstruction experiments, we tune the $r_s$ threshold to ensure effective rejection of outliers and meanwhile to keep as many relative motions as possible.


\subsection{Joint Observations}
\begin{figure} [t]
    \centering
    \includegraphics[width=0.85\linewidth]{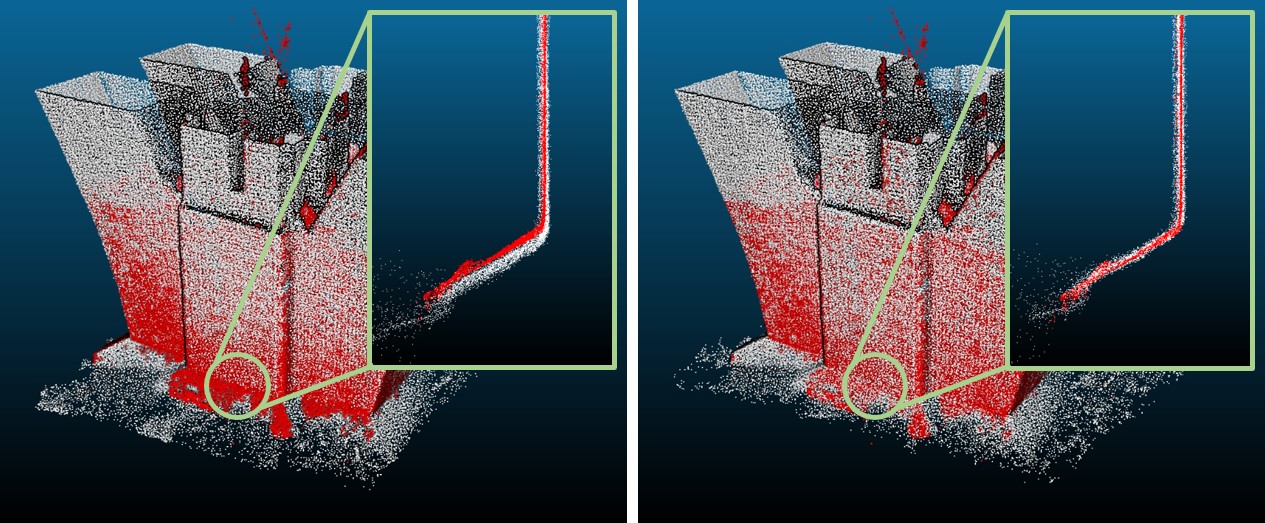}
    \caption{Overlay of LiDAR point cloud (grey) with reconstructed visual features (red). \emph{Left:} Without joint observations. \emph{Right:} With joint observations.}\vspace{-2mm}
    \label{fig:3doverlay}
\end{figure}

For simplicity, we use a smaller scale dataset, namely the bridge pillar dataset, to show the benefits of modeling joint observations in the joint optimization. As shown in Fig. \ref{fig:3doverlay}, the structures estimated from images are overlaid with the merged LiDAR point cloud. From the zoom-in view, it is easy to see that including the joint observations leads to better alignment of the two models, indicating more accurate $\bold T_e$. This observation can be explained by analyzing the observability of $\bold T_e$ discussed in \cite{tsai1989new} and \cite{zhen2019joint}. Basically, if no joint observations are used, $\bold T_e$ is constrained by the trajectories of two sensors. The solution may not be unique if the two trajectories are degraded. An extreme case is the straight-line trajectories. In this case, the translation of $\bold T_e$ could take arbitrary values, while still satisfies the rigid body transform constraints between two trajectories. On the other hand, joint observations are directly shared by the camera and the LiDAR, hence better constrains $\bold T_e$. 

\subsection{Reconstructions}
The reconstruction results on collected datasets are shown in Fig. \ref{fig:smith_recon}. In the first row, we show the reconstruction of small concrete structures. The second row compares the Smith Hall reconstruction results using COLMAP, OpenMVG, and our pipeline. In the three tests, left and right images are used for reconstruction. However, both COLMAP and OpenMVG fail to handle the visual ambiguities caused by the parking signs (shown in Fig. \ref{fig:validation}) and the limited overlapped images. Therefore the generated models are either inconsistent or incomplete. Applying our pipeline helps to reject the invalid motions effectively and allows for building a more consistent model. We encourage readers to view the supplementary material of this paper for more visualization of the reconstructed models.


\section{Conclusions} 
\label{sec:conclusion}
This paper proposes a LiDAR-enhanced stereo SfM pipeline that uses LiDAR information to enhance the robustness, accuracy, consistency and completeness of the components of stereo SfM. Experiment results show that the proposed method is effective in finding more valid motion pairs and eliminating visual ambiguities. Additionally, we show that incorporating the joint observations of the camera and the LiDAR helps to fully constrain the extrinsic transform. Finally, the LiDAR-enhanced SfM pipeline can produce more consistent reconstruction results than the state-of-the-art SfM methods.

\section{Acknowledge}
This work is supported by the Shimizu Institute of Technology, Tokyo. The authors are thankful to Daisuke Hayashi for helping with the data collection. 

\bibliographystyle{IEEEtran.bst}
\bibliography{root.bbl}

\end{document}